\pdfoutput=1
\documentclass[11pt]{article}

\usepackage[]{acl}

\usepackage{times}
\usepackage{latexsym}
\usepackage{graphicx}
\usepackage{amsmath}
\usepackage{multirow}
\usepackage{subcaption}
\usepackage[T1]{fontenc}
\usepackage[utf8]{inputenc}

\usepackage{microtype}

\usepackage{inconsolata}

\title{Improving Acoustic Word Embeddings through Correspondence Training of Self-supervised Speech Representations}

\author{Amit Meghanani, Thomas Hain \\
        Speech and Hearing Research Group \\ Department of Computer Science, The University of Sheffield, United Kingdom \\ \texttt{\{ameghanani1,t.hain\}@sheffield.ac.uk} }

\begin{document}
\maketitle
\begin{abstract}
Acoustic word embeddings (AWEs) are vector representations of spoken words. An effective method for obtaining AWEs is the Correspondence Auto-Encoder (CAE). In the past, the CAE method has been associated with traditional MFCC features. Representations obtained from self-supervised learning (SSL)-based speech models such as HuBERT, Wav2vec2, etc., are outperforming MFCC in many downstream tasks. However, they have not been well studied in the context of learning AWEs. This work explores the effectiveness of CAE with SSL-based speech representations to obtain improved AWEs. Additionally, the capabilities of SSL-based speech models are explored in cross-lingual scenarios for obtaining AWEs. Experiments are conducted on five languages: Polish, Portuguese, Spanish, French, and English. HuBERT-based CAE model achieves the best results for word discrimination in all languages, despite HuBERT being pre-trained on English only. Also, the HuBERT-based CAE model works well in cross-lingual settings. It outperforms MFCC-based CAE models trained on the target languages when trained on one source language and tested on target languages.

\end{abstract}

\section{Introduction}

\label{sec:1}
Self-supervised learning (SSL)-based speech representations are becoming popular in speech processing and producing state-of-the-art results in many downstream tasks such as automatic speech recognition, speaker verification, keyword spotting, voice conversion, etc \cite{superb}. These representations are obtained using self-supervised learning on large amounts of unlabelled speech data. Wav2vec2 \cite{wav2vec2}, HuBERT \cite{hubert}, and WavLM \cite{wavlm} are a few examples of such SSL-based speech models. However, representations obtained from these models have not been extensively explored in the context of learning acoustic word embeddings (AWEs).  AWEs are fixed-dimensional vector representations of spoken words that find applications in various downstream tasks, such as query-by-example search \cite{QbE2, qbe_interspech18,ASE}, keyword spotting \cite{intro_kws}, providing clues for human lexical processing \cite{awe_iclr}, hate speech detection in low resource settings \cite{hate_speech}, etc. 

Recently, the work \cite{AWE-icassp2023} proposed  extracting AWEs from SSL-based speech representations using a mean pooling mechanism. The authors suggest that SSL-based speech representations, which are contextualized, can be effectively converted into AWEs using a straightforward pooling mechanism. On the other hand,  Correspondence Auto-Encoder (CAE) based training strategies for AWEs  \cite{cae-rnn-1} using MFCC \cite{mfcc1980} features are shown to be promising in the literature. Correspondence training involves an auto-encoder where a spoken word serves as the input to the encoder, and the target output of the decoder is a different instance of the same spoken word. This approach helps to preserve acoustic-phonetic information while filtering out unnecessary details such as speaker, acoustic environment, and duration, etc. Both encoder and decoder are typically recurrent neural networks (RNNs). More details about the model will be presented in Sec. \ref{sec:2} and Sec. \ref{sec4.2}. Correspondence training has also been explored in the work \cite{Amit-score} to improve content representations of SSL-based speech models.

The work \cite{correspondence_transformer} uses a Correspondence Transformer Encoder (CTE) for obtaining robust AWEs, trained from scratch and a large-scale unlabelled speech corpus. In contrast, in this work, pre-trained SSL speech models are coupled with a simple RNN based auto-encoder for correspondence training to obtain robust AWEs. This work attempts to use the correspondence training of auto-encoder to obtain the AWEs by leveraging SSL-based speech representations instead of MFCC features as input features to the CAE model. Further, cross-lingual capabilities are also examined for SSL-based AWEs trained with CAE method. The SSL models (HuBERT, Wav2vec2, and WavLM) used in this work are pre-trained on English data. However, it has been demonstrated that these models work well as feature extractors for the all the languages considered in this study. The performances on the word-discrimination task for all the languages (Polish, Portuguese, Spanish, and French) are as good as on the English language (Sec. \ref{sec:5}). A detailed analysis is also conducted to assess the importance of contextual information in spoken words by comparing feature extraction with and without context. For this work, we obtain spoken words for all five languages from the subset of Multilingual LibriSpeech (MLS) \cite{MLS} dataset. The derived dataset consists of five languages (Polish, Portuguese, Spanish, French, English) with start and end timestamps of spoken words \footnote{\href{https://github.com/Trikaldarshi/SSL_AWE}{https://github.com/Trikaldarshi/SSL\_AWE}}.
We chose MLS dataset for our experiments as many previous works \cite{awe_iclr, abdullah-etal-2021-familiar, abdullah22_interspeech, abdullah21_interspeech,multilingual_awe} on AWEs rely on GobalPhone \cite{globalphone} dataset, which is not freely available.

The main contributions of this work are as follows:
\begin{enumerate}
    \item Utilizing corresponding training with SSL-based speech representations to obtain highly discriminative AWEs.

    \item Showing effectiveness of SSL models, pre-trained only on English, as feature extractors in cross-lingual scenarios for obtaining high-quality AWEs.
    \item Quantitatively demonstrating that incorporating the context of the spoken word in SSL-based speech representations leads to the production of more robust AWEs.
\end{enumerate}

The rest of the paper is as follows: Sec. \ref{sec:2} describes the correspondence auto-encoder methodology to obtain AWEs;  Sec. \ref{sec:3} describes the data preparation and data statistics; Sec. \ref{sec:4} describes the details of the experiments; Sec. \ref{sec:5} describes the results and analysis; Sec. \ref{sec:6} concludes the work with possible future directions.
Sec. \ref{sec:7} describes the limitations of the work.

\section{Methodology}
\label{sec:2}
Correspondence auto-encoder is trained with input as a spoken word and target output as a different instance of the same spoken word. Typically, Recurrent Neural Network (RNN) based encoder and decoder are used, hence the model is referred to as CAE-RNN. The rationale behind this training method is that CAE-RNN will preserve only the acoustic-phonetic information and filter out the other unnecessary information factors such as speaker, duration, acoustic environment, etc \cite{cae-rnn-1}. Fig. \ref{fig:fig1} shows the CAE-RNN model setup. The input to the ENC is a sequence of acoustic feature vectors ($X = {X_1, X_2,..., X_m}$) of a spoken word. The target output is the sequence of acoustic feature vectors of the different instance of the same spoken word ($X' = {X'_1, X'_2,..., X'_n}$). The encoder produces the AWE (e) of the spoken word $X$, which is then fed to the decoder to reconstruct $X'$. The output of the decoder is represented as $Y = {Y_1, Y_2,..., Y_n}$. The mean squared loss function for a single training pair ($X,X'$)  can be described as following:

\begin{figure}
    \centering
    \includegraphics[width=\columnwidth]{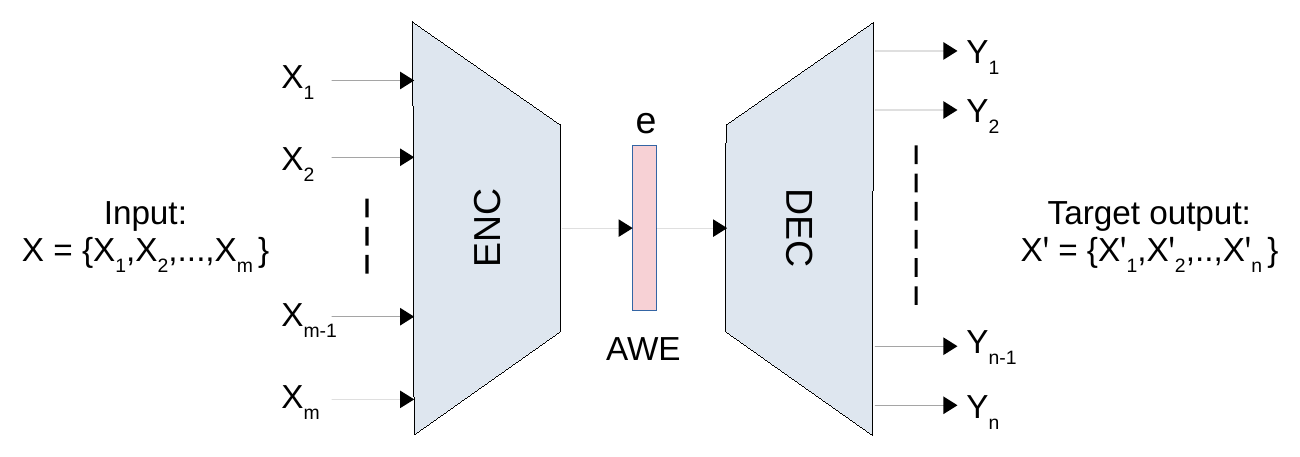}
    \caption{CAE-RNN training setup for extracting AWEs \cite{cae-rnn-1}.}
    \label{fig:fig1}
\end{figure}

\begin{equation}
\label{eq1}
    L = \sum_{k=1}^{n}||X'_{k} - Y_{k}||^2
\end{equation}
, where $X' = {X'_1, X'_2,..., X'_n}$ is the target output and $Y = {Y_1, Y_2,..., Y_n}$ is the output of the decoder as shown in Fig. \ref{fig:fig1}.

\section{Data Preparation}

\label{sec:3}

\begin{table*}[ht]

\resizebox{\textwidth}{!}{%
\setlength{\tabcolsep}{4pt}
\begin{tabular}{cccccccccccccccc}
\hline
\multirow{2}{*}{\textbf{\begin{tabular}[c]{@{}c@{}}Data \\ Statistics\end{tabular}}} & \multicolumn{3}{c}{\textbf{Polish}} & \multicolumn{3}{c}{\textbf{Portuguese}} & \multicolumn{3}{c}{\textbf{Spanish}} & \multicolumn{3}{c}{\textbf{French}} & \multicolumn{3}{c}{\textbf{English}} \\ \cline{2-16} 
                                                                                     & Train       & Dev       & Test      & Train         & Dev        & Test       & Train       & Dev        & Test      & Train       & Dev       & Test      & Train       & Dev        & Test      \\ \hline
\# Spoken Words                                                                      & 104448      & 4595      & 4563      & 117820        & 4964       & 4659       & 82258       & 3721       & 3601      & 84267       & 3004      & 3114      & 92352       & 3147       & 3192      \\
\# Unique Spoken Words                                                               & 9346        & 3818      & 3887      & 9785          & 3696       & 3539       & 7085        & 2678       & 2769      & 7221        & 2394      & 2433      & 7157        & 2448       & 2527      \\
\# Speaker                                                                           & 11          & 4         & 4         & 42            & 10         & 10         & 79          & 20         & 20        & 120         & 18        & 18        & 182         & 42         & 41        \\
Total Duration (hours)                                                               & 18.5        & 0.8       & 0.8       & 21.9          & 0.93       & 0.87       & 14.7        & 0.67       & 0.65      & 14.9        & 0.55      & 0.56      & 16.6        & 0.56       & 0.57      \\ \hline
\end{tabular}%
}
\caption{A summary of the data statistics for all five languages across the train, dev, and test splits}
\label{tab1}
\end{table*}

The Multilingual LibriSpeech (MLS) dataset \cite{MLS} is utilized  to obtain spoken words. Five languages, namely Polish, Portuguese, Spanish, French, and English, are selected from MLS. For each language, approximately 25,000 utterances are selected for the training set, 500 for the development set, and 500 for the test set. These selected utterances are force-aligned to obtain the spoken word boundaries using the Montreal Forced Aligner toolkit \cite{mfa}. Only spoken words with a duration of 0.5 seconds or longer are included in the derived dataset, following the standard practice in the literature \cite{multiview}. Spoken words with a frequency greater than 50 or less than 5 are excluded from the derived dataset. Table \ref{tab1} presents a summary of the statistics for the final extracted dataset
, encompassing all five languages. The speakers across different sets are non-overlapping, which is a desirable characteristic for evaluating AWEs as they should be robust to speaker variations. Polish language had limited available data and consequently has the fewest number of speakers, while English has the highest number. The duration in Table \ref{tab1} represents the total time duration of spoken words across the different sets.
\section{Experimental Setup}
Experiments are conducted on all the five languages with SSL-based speech representations as input features extracted from Wav2vec2, HuBERT, and WavLM. Experiments are also conducted with MFCC as input features.
First, the feature extraction methods for various SSL-based speech representations and MFCCs are described. Then, the configuration of the CAE-RNN model is explained, along with the mean pooling baseline \cite{AWE-icassp2023} and the AE-RNN method (without correspondence training), for comparison. Next, the word discrimination task is described, which is used for evaluating the quality of the extracted AWEs. Finally, the training details of the CAE-RNN and other models are provided.
\label{sec:4}
\subsection{Feature Extraction}

\subsubsection{SSL-based Speech Representations}
SSL models are pre-trained on large amount of unlabelled speech data. The task defined for the pre-training is known as the pretext task. Each model differs based on how the pretext task was defined, the data used for pre-training, and the model architecture. In this work, the ``BASE'' architectures of Wav2vec2, HuBERT, and WavLM model \footnote{\href{https://github.com/pytorch/fairseq}{https://github.com/pytorch/fairseq}} (all with $\approx$ 95M parameters) are used for feature extraction. All these models are pre-trained on 960 hours of LibriSpeech data \cite{libri}. A ``BASE" architecture typically has a multi-layer CNN-based feature encoder followed by 12 Transformer layers. In this work, representations from each model are extracted from the final (i.e. 12\textsuperscript{th}) Transformer layer. For all the above mentioned SSL models, 768-dimensional feature vectors are obtained for each spoken word at a framerate of 20 ms.

SSL-based speech representations are extracted in two different ways: the first one is extracted using the context around the spoken word, and the other one is extracted without the context, as described here:
\begin{enumerate}
\item \textbf{With context:} In this case, first the SSL-based speech representations of the entire spoken utterance are computed and then the  time boundaries of the spoken word is used to get the representations of the segment belonging to the spoken word. This ensures that the extracted representations capture the context around the spoken word as the entire utterance is processed by the SSL model.  Let us assume $U$ represents an utterance and $X$ represents a spoken word instance present in the utterance $U$ with start and end timestamps denoted as $t_1$ and $t_2$. If $f$ represents the SSL model, then the SSL-based speech representation for the entire utterance is computed $Z = f(U)$. Then the speech representations for the spoken word $X$ will be $Z_{t_1:t_2}$.

\item \textbf{Without context:} In this case, no context is considered and SSL-based speech representations are extracted by inputting only speech segments belonging to the spoken words to the SSL models. Hence, in this case, the SSL-based speech representation for the spoken word $X$ will be  $Z = f(U_{t_1:t_2})$.
\end{enumerate}

\subsubsection{MFCC Features}
For each spoken word, 20-dimensional MFCC features are extracted with 30 ms window size and 20 ms shift along with delta and delta-delta features, which leads to 60-dimensional MFCC feature vectors.
\subsection{Model Details}
\label{sec4.2}
A 4-layer Bidirectional GRU with a hidden dimension of 256 is used for both the encoder and decoder in the CAE-RNN model. Dropout rate is set to 0.2. The final hidden state of the encoder-GRU is fed to a fully connected layer to obtain 128-dimensional AWE (e) as shown in Fig. \ref{fig:fig1}. This embedding is then fed to the decoder at each time step as input to the decoder \cite{cae-rnn-1}. The output of the decoder is then fed to a fully connected layer to produce the target output.

A regular auto-encoder RNN (AE-RNN) model is also used as one of the baselines with similar configurations. AE-RNN model is an auto-encoder model where input and target output is exactly the same spoken word, i.e. input-output training pair is ($X,X$). A mean pooling model is also used as baseline \cite{AWE-icassp2023}, which does not require any training. This method computes the mean of the SSL-based speech representations to get the 768-dimensional AWE of a spoken word.

\subsection{Word Discrimination Task}
\label{sec4.3}
To evaluate the AWEs, the same-different word-discrimination task is used \cite{ap1,first_ap}. First, all possible spoken word pairs are generated. For example, if there are total N spoken words, then the total generated spoken word pairs for comparison will be $ \binom{N}{2} = \frac{N(N-1)}{2} $. After that, the cosine distance between the AWEs of these pairs are computed and compared with a threshold to decide whether the spoken words are same or different. The average precision (AP) is calculated by varying all the possible threshold values, which is the area under the precision-recall curve. AP is reported for the same-different word discrimination task. 
Word-discrimination task is applied on the test set, which has unseen speakers during training.
Also, a subset of the test set is created for all five languages in such a way that none of the words in the subset are encountered during training. This particular subset is referred to as \textbf{$\text{test}'$}. The word-discrimination task is also conducted on $\text{test}'$. The total number of generated spoken word pairs for both the test and $\text{test}'$ sets is described in Table \ref{tab-word-par} for all languages.

\begin{table}[]

\resizebox{\columnwidth}{!}{%
\setlength{\tabcolsep}{3pt}
\begin{tabular}{cccccc}
\hline
\textbf{\begin{tabular}[c]{@{}c@{}}Spoken Word\\ Pairs (in million)\end{tabular}} & \textbf{Polish} & \textbf{Portuguese} & \textbf{Spanish} & \textbf{French} & \textbf{English} \\ \hline
\textbf{test}                                                                     & 10.4 M          & 10.8 M              & 6.4 M            & 4.8 M           & 5.1 M            \\
\hspace{0.025cm} \textbf{$\text{test}'$}                                                                    & 4.7             & 2.6 M               & 1.8 M            & 1.3 M           & 1.4 M      \\
\hline
\end{tabular}%
}
\caption{The total number of spoken word pairs generated for the \textbf{test} and \textbf{$\text{test}'$} sets.}
\label{tab-word-par}
\end{table}

\subsection{Training Details} 
The total number of generated correspondence training pairs ($X,X'$) for each language is as follows: 9,55,106 for Polish, 11,63,468 for Portuguese, 7,80,197 for Spanish, 7,95,613 for French, and 9,72,532 for English. The remaining training details are as follows for various inputs: 
\begin{itemize}
    \item \textbf{SSL-based Speech Representations as Input:}
CAE-RNN models are trained for 30 epochs, using a learning rate of 0.0001 with Adam optimizer and a batch size of 512. In each run, the model with the best performance on the development set in terms of word-discrimination is selected as the final model for evaluation on the test set. AE-RNN models are trained for 50 epochs, keeping all other parameters same as mentioned above for the CAE-RNN models.

\item \textbf{MFCC as Input:} Both AE-RNN and CAE-RNN models with MFCC as inputs are trained for 100 epochs, using a learning rate of 0.0001 with Adam optimizer. The batch size for the AE-RNN model was chosen as 64, while for the CAE-RNN model it was set to 256 based on preliminary experiments for better convergence. Similarly to the previous case, the model with the best performance on the development set in terms of word-discrimination is selected as the final model for evaluation on the test set.

\end{itemize}

\begin{table}[ht]
\centering

\resizebox{\columnwidth}{!}{%
\setlength{\tabcolsep}{3pt}
\begin{tabular}{cccccc}
\hline
\textbf{Model}   & \textbf{Polish} & \textbf{Portuguese} & \textbf{Spanish} & \textbf{French} & \textbf{English} \\ \hline
AE-RNN  & 0.20            & 0.10                & 0.17             & 0.01            & 0.01             \\ 
CAE-RNN & 0.56            & 0.41                & 0.57             & 0.43            & 0.24             \\ \hline
\end{tabular}%
}
\caption{AP on the \textbf{test} set for word-discrimination task using MFCCs as input features for AE-RNN and CAE-RNN models in different languages.}
\label{tab2}
\end{table}

\begin{table}[ht]
\centering

\resizebox{\columnwidth}{!}{%
\setlength{\tabcolsep}{3pt}
\begin{tabular}{cccccc}
\hline
\textbf{Model}   & \textbf{Polish} & \textbf{Portuguese} & \textbf{Spanish} & \textbf{French} & \textbf{English} \\ \hline
AE-RNN  & 0.21            & 0.10                & 0.24             & 0.03            & 0.01             \\ 
CAE-RNN & 0.54            & 0.47                & 0.63             & 0.57            & 0.33             \\ \hline
\end{tabular}%
}
\caption{AP on the \textbf{$\text{test}'$} set for word-discrimination task using MFCCs as input features for AE-RNN and CAE-RNN models in different languages.}
\label{tab3}
\end{table}

\section{Results and Analysis}

\begin{table*}[ht]
\centering

\resizebox{\textwidth}{!}{%
\setlength{\tabcolsep}{5pt}
\begin{tabular}{cccccccccccc}
\hline
\multirow{2}{*}{\textbf{\begin{tabular}[c]{@{}c@{}}AWE\\ Extraction\\ Method\end{tabular}}} &
  \multirow{2}{*}{\textbf{\begin{tabular}[c]{@{}c@{}}Input\\ Features\end{tabular}}} &
  \multicolumn{2}{c}{\textbf{Polish}} &
  \multicolumn{2}{c}{\textbf{Portuguese}} &
  \multicolumn{2}{c}{\textbf{Spanish}} &
  \multicolumn{2}{c}{\textbf{French}} &
  \multicolumn{2}{c}{\textbf{English}} \\ \cline{3-12} 
 &
   &
  \multicolumn{1}{c}{\textbf{\begin{tabular}[c]{@{}c@{}}with\\ context\end{tabular}}} &
  \textbf{\begin{tabular}[c]{@{}c@{}}without\\ context\end{tabular}} &
  \multicolumn{1}{c}{\textbf{\begin{tabular}[c]{@{}c@{}}with\\ context\end{tabular}}} &
  \textbf{\begin{tabular}[c]{@{}c@{}}without\\ context\end{tabular}} &
  \multicolumn{1}{c}{\textbf{\begin{tabular}[c]{@{}c@{}}with\\ context\end{tabular}}} &
  \textbf{\begin{tabular}[c]{@{}c@{}}without\\ context\end{tabular}} &
  \multicolumn{1}{c}{\textbf{\begin{tabular}[c]{@{}c@{}}with\\ context\end{tabular}}} &
  \textbf{\begin{tabular}[c]{@{}c@{}}without\\ context\end{tabular}} &
  \multicolumn{1}{c}{\textbf{\begin{tabular}[c]{@{}c@{}}with\\ context\end{tabular}}} &
  \textbf{\begin{tabular}[c]{@{}c@{}}without\\ context\end{tabular}} \\ \hline
\multirow{3}{*}{Mean Pooling} &
  Wav2vec2 &
  \multicolumn{1}{c}{0.01} &
  0.00 &
  \multicolumn{1}{c}{0.00} &
  0.00 &
  \multicolumn{1}{c}{0.02} &
  0.00 &
  \multicolumn{1}{c}{0.03} &
  0.01 &
  \multicolumn{1}{c}{0.02} &
  0.00 \\
 &
  WavLM &
  \multicolumn{1}{c}{0.07} &
  0.00 &
  \multicolumn{1}{c}{0.01} &
  0.00 &
  \multicolumn{1}{c}{0.03} &
  0.00 &
  \multicolumn{1}{c}{0.05} &
  0.02 &
  \multicolumn{1}{c}{0.07} &
  0.03 \\
 &
  HuBERT &
  \multicolumn{1}{c}{0.17} &
  0.33 &
  \multicolumn{1}{c}{0.10} &
  0.15 &
  \multicolumn{1}{c}{0.14} &
  0.32 &
  \multicolumn{1}{c}{0.22} &
  0.31 &
  \multicolumn{1}{c}{0.15} &
  0.24 \\ \hline
\multirow{3}{*}{AE-RNN} &
  Wav2vec2 &
  \multicolumn{1}{c}{0.10} &
  0.00 &
  \multicolumn{1}{c}{0.08} &
  0.00 &
  \multicolumn{1}{c}{0.15} &
  0.01 &
  \multicolumn{1}{c}{0.13} &
  0.01 &
  \multicolumn{1}{c}{0.07} &
  0.00 \\ 
 &
  WavLM &
  \multicolumn{1}{c}{0.34} &
  0.14 &
  \multicolumn{1}{c}{0.21} &
  0.11 &
  \multicolumn{1}{c}{0.43} &
  0.20 &
  \multicolumn{1}{c}{0.34} &
  0.23 &
  \multicolumn{1}{c}{0.26} &
  0.17 \\ 
 &
  HuBERT &
  \multicolumn{1}{c}{0.44} &
  0.40 &
  \multicolumn{1}{c}{0.36} &
  0.27 &
  \multicolumn{1}{c}{0.58} &
  0.52 &
  \multicolumn{1}{c}{0.45} &
  0.40 &
  \multicolumn{1}{c}{0.36} &
  0.34 \\ \hline
\multirow{3}{*}{CAE-RNN} &
  Wav2vec2 &
  \multicolumn{1}{c}{0.86} &
  0.71 &
  \multicolumn{1}{c}{0.86} &
  0.63 &
  \multicolumn{1}{c}{0.93} &
  0.79 &
  \multicolumn{1}{c}{0.71} &
  0.61 &
  \multicolumn{1}{c}{0.82} &
  0.52 \\ 
 &
  WavLM &
  \multicolumn{1}{c}{0.86} &
  0.72 &
  \multicolumn{1}{c}{0.76} &
  0.63 &
  \multicolumn{1}{c}{0.92} &
  0.85 &
  \multicolumn{1}{c}{0.70} &
  0.61 &
  \multicolumn{1}{c}{0.66} &
  0.51 \\ 
 &
  HuBERT &
  \multicolumn{1}{c}{\textbf{0.90}} &
  \textbf{0.82} &
  \multicolumn{1}{c}{\textbf{0.88}} &
  \textbf{0.71} &
  \multicolumn{1}{c}{\textbf{0.95}} &
  \textbf{0.89} &
  \multicolumn{1}{c}{\textbf{0.74}} &
  \textbf{0.65} &
  \multicolumn{1}{c}{\textbf{0.86}} &
  \textbf{0.65} \\ \hline
\end{tabular}%
}
\caption{AP scores for the word-discrimination task on the \textbf{test} set using SSL-based speech representations as input features for all five languages. AWE extraction methods include mean pooling \cite{AWE-icassp2023}, AE-RNN, and CAE-RNN.}
\label{tab4}
\end{table*}

Table \ref{tab2} shows the baseline results with MFCC features as input for the AE-RNN and CAE-RNN models. This demonstrates the effectiveness of the CAE-RNN model over the AE-RNN model for the word-discrimination task, as the CAE-RNN consistently outperforms the AE-RNN for all languages. Table \ref{tab3} presents the results for the derived subset of the test set ($\text{test}'$) with similar trends. It is worth noting that the AP on the $\text{test}'$ set is relatively better than that of the original test set in most cases. This is likely due to the fact that the number of spoken word pairs generated for the evaluation on the $\text{test}'$ is significantly fewer compared to the original test set, as mentioned in Table \ref{tab-word-par}.

Table \ref{tab4} displays the results obtained from using various SSL-based speech representations (Wav2vec2, WavLM, and HuBERT) as input features, combined with different AWE extraction methods (mean pooling, CAE-RNN, and AE-RNN). The results presented in Table \ref{tab4} represent the AP for the word-discrimination task on the test set, employing different SSL-based speech representation feature extraction setups (`with context' and `without context').
From Table \ref{tab4}, it is evident that the AWEs derived `with context' exhibit greater robustness. The AP on the test set for all languages is significantly better when utilizing SSL-based speech representations `with context' compared to the feature extraction `without context'.
As shown in Table \ref{tab4}, the CAE-RNN model demonstrates superior performance when using SSL-based speech representations as input features compared to the MFCC-based baseline model (Table \ref{tab2}) across all languages. Furthermore, Table \ref{tab4} provides a comparison of the CAE-RNN model with other baseline models (mean pooling and AE-RNN) when utilizing SSL-based speech representations as input features. CAE-RNN consistently outperforms both the AE-RNN and mean pooling methods for all languages and SSL models. Another advantage of the CAE-RNN model over mean pooling is that the AWEs obtained from CAE-RNN have a dimension of 128, while mean pooling-based AWEs are 768-dimensional. Based on the results presented in Table \ref{tab4}, it is evident that the HuBERT features consistently achieve the best performance across all configurations and languages. Specifically, when using the CAE-RNN method for AWE extraction and SSL-based speech representations extracted `with context', the HuBERT achieves the highest AP on the test set: 0.90 for Polish, 0.88 for Portuguese, 0.95 for Spanish, 0.74 for French, and 0.86 for English. The performance order can be sorted as HuBERT $>$ Wav2vec2 $>$ WavLM $>$ MFCCs when using the CAE-RNN-based AWE model and SSL-based speech representations extracted `with context'.

Table \ref{tab5} presents the results for the $\textbf{test}'$ set, which includes unseen words and speakers. The models exhibit similar trends in performance in this case as well. This provides evidence that the proposed methodology performs equally well on unseen words. 
One interesting finding is that the SSL-based speech representations considered in this work were pre-trained solely on English language. Despite this, they are capable of generating meaningful features for other languages, resulting in good performance as demonstrated in Table \ref{tab4} and \ref{tab5} for the word-discrimination task.

\label{sec:5}
\begin{table*}[ht]
\centering

\resizebox{\textwidth}{!}{%
\setlength{\tabcolsep}{5pt}
\begin{tabular}{cccccccccccc}
\hline
\multirow{2}{*}{\textbf{\begin{tabular}[c]{@{}c@{}}AWE\\ Extraction\\ Method\end{tabular}}} &
  \multirow{2}{*}{\textbf{\begin{tabular}[c]{@{}c@{}}Input\\ Features\end{tabular}}} &
  \multicolumn{2}{c}{\textbf{Polish}} &
  \multicolumn{2}{c}{\textbf{Portuguese}} &
  \multicolumn{2}{c}{\textbf{Spanish}} &
  \multicolumn{2}{c}{\textbf{French}} &
  \multicolumn{2}{c}{\textbf{English}} \\ \cline{3-12} 
 &
   &
  \multicolumn{1}{c}{\textbf{\begin{tabular}[c]{@{}c@{}}with\\ context\end{tabular}}} &
  \textbf{\begin{tabular}[c]{@{}c@{}}without\\ context\end{tabular}} &
  \multicolumn{1}{c}{\textbf{\begin{tabular}[c]{@{}c@{}}with\\ context\end{tabular}}} &
  \textbf{\begin{tabular}[c]{@{}c@{}}without\\ context\end{tabular}} &
  \multicolumn{1}{c}{\textbf{\begin{tabular}[c]{@{}c@{}}with\\ context\end{tabular}}} &
  \textbf{\begin{tabular}[c]{@{}c@{}}without\\ context\end{tabular}} &
  \multicolumn{1}{c}{\textbf{\begin{tabular}[c]{@{}c@{}}with\\ context\end{tabular}}} &
  \textbf{\begin{tabular}[c]{@{}c@{}}without\\ context\end{tabular}} &
  \multicolumn{1}{c}{\textbf{\begin{tabular}[c]{@{}c@{}}with\\ context\end{tabular}}} &
  \textbf{\begin{tabular}[c]{@{}c@{}}without\\ context\end{tabular}} \\ \hline
\multirow{3}{*}{Mean Pooling} &
  Wav2vec2 &
  \multicolumn{1}{c}{0.01} &
  0.01 &
  \multicolumn{1}{c}{0.01} &
  0.00 &
  \multicolumn{1}{c}{0.04} &
  0.01 &
  \multicolumn{1}{c}{0.06} &
  0.01 &
  \multicolumn{1}{c}{0.03} &
  0.00 \\ 
 &
  WavLM &
  \multicolumn{1}{c}{0.05} &
  0.00 &
  \multicolumn{1}{c}{0.02} &
  0.01 &
  \multicolumn{1}{c}{0.05} &
  0.01 &
  \multicolumn{1}{c}{0.10} &
  0.05 &
  \multicolumn{1}{c}{0.09} &
  0.04 \\ 
 &
  HuBERT &
  \multicolumn{1}{c}{0.18} &
  0.31 &
  \multicolumn{1}{c}{0.11} &
  0.17 &
  \multicolumn{1}{c}{0.19} &
  0.40 &
  \multicolumn{1}{c}{0.30} &
  0.40 &
  \multicolumn{1}{c}{0.18} &
  0.29 \\ \hline
\multirow{3}{*}{AE-RNN} &
  Wav2vec2 &
  \multicolumn{1}{c}{0.09} &
  0.01 &
  \multicolumn{1}{c}{0.09} &
  0.01 &
  \multicolumn{1}{c}{0.24} &
  0.02 &
  \multicolumn{1}{c}{0.21} &
  0.03 &
  \multicolumn{1}{c}{0.08} &
  0.01 \\
 &
  WavLM &
  \multicolumn{1}{c}{0.33} &
  0.12 &
  \multicolumn{1}{c}{0.25} &
  0.12 &
  \multicolumn{1}{c}{0.55} &
  0.28 &
  \multicolumn{1}{c}{0.44} &
  0.33 &
  \multicolumn{1}{c}{0.29} &
  0.23 \\ 
 &
  HuBERT &
  \multicolumn{1}{c}{0.44} &
  0.41 &
  \multicolumn{1}{c}{0.41} &
  0.31 &
  \multicolumn{1}{c}{0.69} &
  0.64 &
  \multicolumn{1}{c}{0.55} &
  0.50 &
  \multicolumn{1}{c}{0.43} &
  0.41 \\ \hline
\multirow{3}{*}{CAE-RNN} &
  Wav2vec2 &
  \multicolumn{1}{c}{0.87} &
  0.72 &
  \multicolumn{1}{c}{0.90} &
  0.65 &
  \multicolumn{1}{c}{0.96} &
  0.85 &
  \multicolumn{1}{c}{0.84} &
  0.73 &
  \multicolumn{1}{c}{0.89} &
  0.67 \\ 
 &
  WavLM &
  \multicolumn{1}{c}{0.85} &
  0.72 &
  \multicolumn{1}{c}{0.81} &
  0.68 &
  \multicolumn{1}{c}{0.95} &
  0.90 &
  \multicolumn{1}{c}{0.83} &
  0.75 &
  \multicolumn{1}{c}{0.75} &
  0.62 \\ 
 &
  HuBERT &
  \multicolumn{1}{c}{\textbf{0.90}} &
  \textbf{0.83} &
  \multicolumn{1}{c}{\textbf{0.91}} &
  \textbf{0.75} &
  \multicolumn{1}{c}{\textbf{0.97}} &
  \textbf{0.93} &
  \multicolumn{1}{c}{\textbf{0.86}} &
  \textbf{0.81} &
  \multicolumn{1}{c}{\textbf{0.93}} &
  \textbf{0.75} \\ \hline
\end{tabular}%
}
\caption{AP scores for the word-discrimination task on the \textbf{$\text{test}'$} set using SSL-based speech representations as input features for all five languages. AWE extraction methods include mean pooling \cite{AWE-icassp2023}, AE-RNN, and CAE-RNN..}

\label{tab5}
\end{table*}

\subsection{Cross-lingual Analysis}
To assess the effectiveness of SSL speech representation-based CAE-RNN models in cross-lingual settings, a CAE-RNN model trained on one source language (English in this case) is evaluated on four different target languages. This evaluation can be considered a `zero-shot' evaluation, as no training data from the target languages is required. Table \ref{tab6} displays the results in terms of AP for the word-discrimination task on the test set and $\text{test}'$ set for the four target languages (Polish, Portuguese, Spanish, and French).
In this scenario as well, the HuBERT-based CAE-RNN model achieves the best performance across all languages, except for French where Wav2vec2 performs the best. It is worth noting that the CAE-RNN model in the `zero-shot' setting outperforms the mean pooling method (Table \ref{tab4} and \ref{tab5}) \cite{AWE-icassp2023} and the CAE-RNN model trained on the target language with MFCC features (Table \ref{tab2} and \ref{tab3}).
The mean pooling method \cite{AWE-icassp2023} can be considered a `zero-shot' AWE extraction method, as it does not involve additional training on top of the pre-trained SSL models. In a `zero-shot' setup for target languages, using a CAE-RNN trained on a well-resourced source language can offer an advantage over the mean pooling method. In conclusion, SSL-based CAE-RNN models have fairly good performance when used crosslingually. There have been earlier studies \cite{multilingual_awe} on acoustic word embeddings for zero-resource languages using multilingual transfer with MFCC features, which worked well. Also, intuitively, some generalisation was expected as the aim of modelling is to compress a small segment of speech into a fixed dimensional vector. There might be a language effect on pre-trained SSL speech models but the basic speech properties are still invariant to changes in the language. The cross-lingual ability of SSL-based CAE-RNN models to obtain AWEs can support many applications such as speech search, indexing and discovery systems for languages with low-resources \cite{multilingual_awe}. 
\begin{table}[ht]
\centering

\resizebox{\columnwidth}{!}{%
\setlength{\tabcolsep}{4pt}
\begin{tabular}{ccccccccc}
\hline
\multirow{3}{*}{\textbf{\begin{tabular}[c]{@{}c@{}}Input \\ Features\end{tabular}}} &
  \multicolumn{2}{c}{\textbf{Polish}} &
  \multicolumn{2}{c}{\textbf{Portuguese}} &
  \multicolumn{2}{c}{\textbf{Spanish}} &
  \multicolumn{2}{c}{\textbf{French}} \\ \cline{2-9} 
 &
  \multicolumn{1}{c}{\multirow{2}{*}{\textbf{test}}} &
  \multirow{2}{*}{\textbf{$\text{test}'$}} &
  \multicolumn{1}{c}{\multirow{2}{*}{\textbf{test}}} &
  \multirow{2}{*}{\textbf{$\text{test}'$}} &
  \multicolumn{1}{c}{\multirow{2}{*}{\textbf{test}}} &
  \multirow{2}{*}{\textbf{$\text{test}'$}} &
  \multicolumn{1}{c}{\multirow{2}{*}{\textbf{test}}} &
  \multirow{2}{*}{\textbf{$\text{test}'$}} \\
         & \multicolumn{1}{c}{}     &      & \multicolumn{1}{c}{}     &      & \multicolumn{1}{c}{}     &      & \multicolumn{1}{c}{}              &               \\ \hline
Wav2vec2 & \multicolumn{1}{c}{0.57} & 0.57 & \multicolumn{1}{c}{0.48} & 0.54 & \multicolumn{1}{c}{0.60} & 0.68 & \multicolumn{1}{c}{\textbf{0.52}} & \textbf{0.68} \\ 
WavLM    & \multicolumn{1}{c}{0.48} & 0.47 & \multicolumn{1}{c}{0.36} & 0.40 & \multicolumn{1}{c}{0.54} & 0.63 & \multicolumn{1}{c}{0.50}          & 0.64          \\ 
HuBERT &
  \multicolumn{1}{c}{\textbf{0.59}} &
  \textbf{0.60} &
  \multicolumn{1}{c}{\textbf{0.50}} &
  \textbf{0.56} &
  \multicolumn{1}{c}{\textbf{0.62}} &
  \textbf{0.69} &
  \multicolumn{1}{c}{0.48} &
  0.65 \\ 
MFCC     & \multicolumn{1}{c}{0.18} & 0.20 & \multicolumn{1}{c}{0.11} & 0.15 & \multicolumn{1}{c}{0.22} & 0.29 & \multicolumn{1}{c}{0.22}          & 0.35          \\ \hline
\end{tabular}%
}
\caption{AP for the word-discrimination task with CAE-RNN model trained on English language  with various input features and tested on other four languages.}
\label{tab6}
\end{table}

\begin{table}[ht]
\centering

\resizebox{\columnwidth}{!}{%
\setlength{\tabcolsep}{3pt}
\begin{tabular}{ccccc}
\hline
\textbf{Word1} &
  \textbf{Word2} &
  \textbf{\begin{tabular}[c]{@{}c@{}}Cosine Distance\\ (Mean Pooling)\end{tabular}} &
  \textbf{\begin{tabular}[c]{@{}c@{}}Cosine Distance\\ (CAE-RNN)\end{tabular}} &
  \textbf{Description} \\ \hline
% aside & aisde &  0.19 & 0.01 & SS-SWP   \\ \hline
aside & aside &  0.23 & 0.01 & Same word    \\ 
aside & ideas & 0.56 & 0.99 & Anagram pair \\ 
% aside & ideas & 0.67  & 1.00  & DS-AWP \\ \hline
this  & this  &  0.33 & 0.11 & Same word         \\ 
this  & hits  & 0.38  & 0.50 & Anagram pair \\ 
% this  & hits  & 0.58 & 0.46 &  DS-AWP \\ \hline
no    & no    & 0.30 & 0.02 & Same word        \\ 
no    & on   & 0.63 & 0.69 &  Anagram pair \\ \hline
% no    & on    & 0.54 & 0.69 &  DS-AWP \\ \hline
\end{tabular}%
}
\caption{{Comparison of cosine distances between AWEs of same
spoken word pairs and anagram pairs. HuBERT features are
used for both CAE-RNN and mean pooling method.}}
\label{tab7}
\end{table}
  % htbp

\begin{figure}[htbp]
  \centering
  
  \begin{subfigure}[b]{\columnwidth}
    \centering
    \includegraphics[width=\columnwidth]{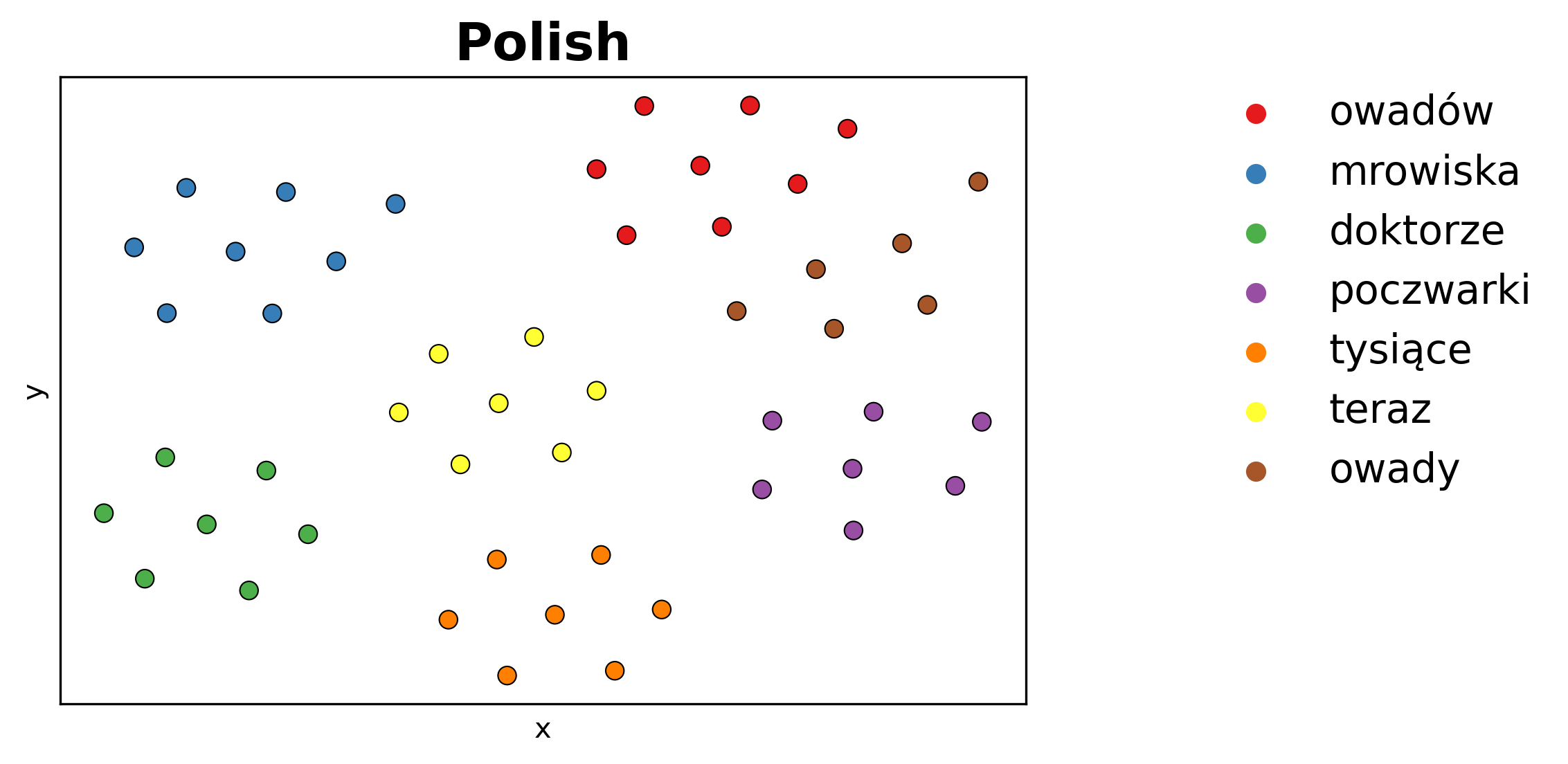}
    % \caption{Figure A}
    \label{fig:a}
  \end{subfigure}
  
  \begin{subfigure}[b]{\columnwidth}
    \centering
    \includegraphics[width=\columnwidth]{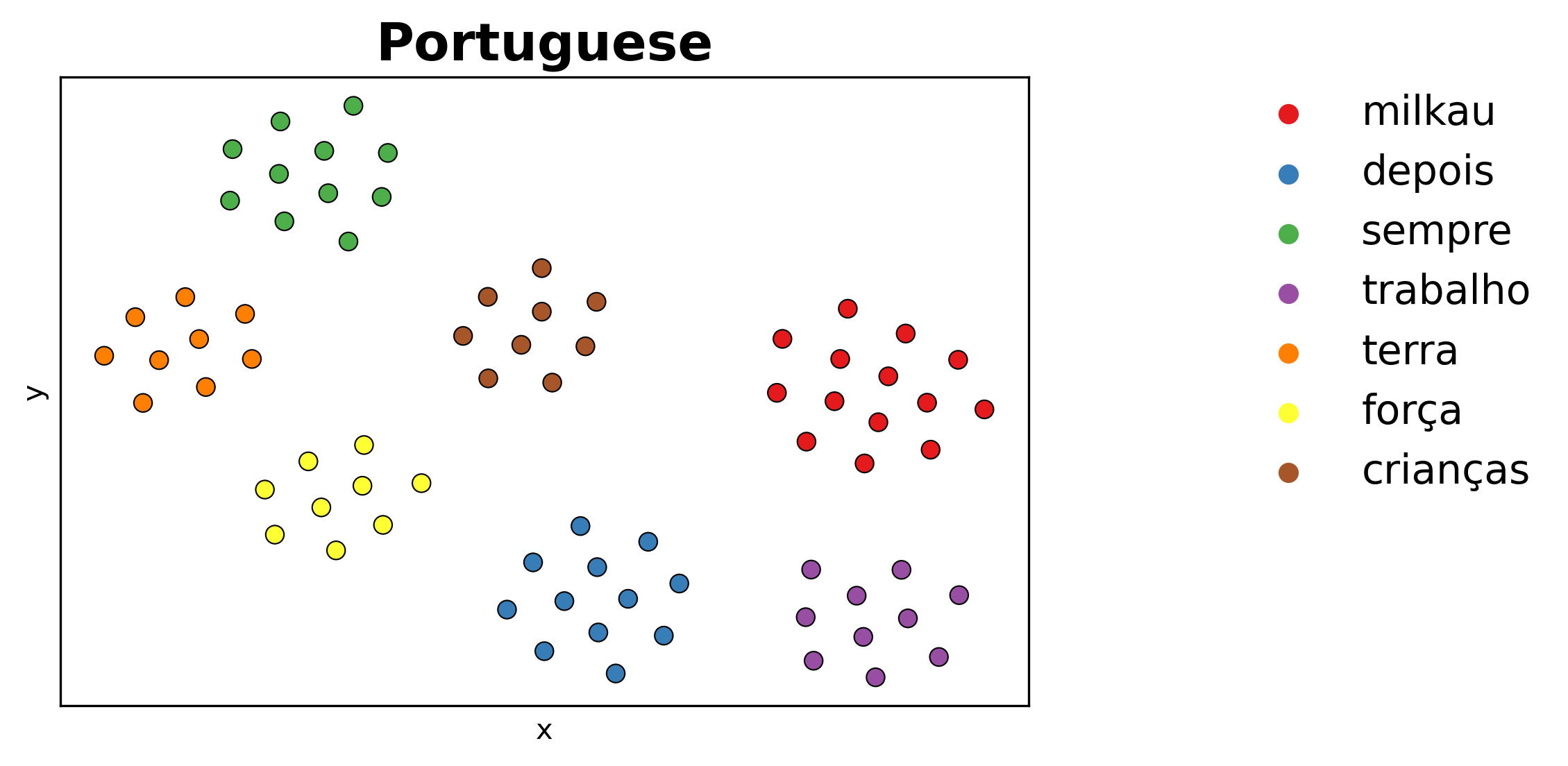}
    % \caption{Figure B}
    \label{fig:b}
  \end{subfigure}
  
  \begin{subfigure}[b]{\columnwidth}
    \centering
    \includegraphics[width=\columnwidth]{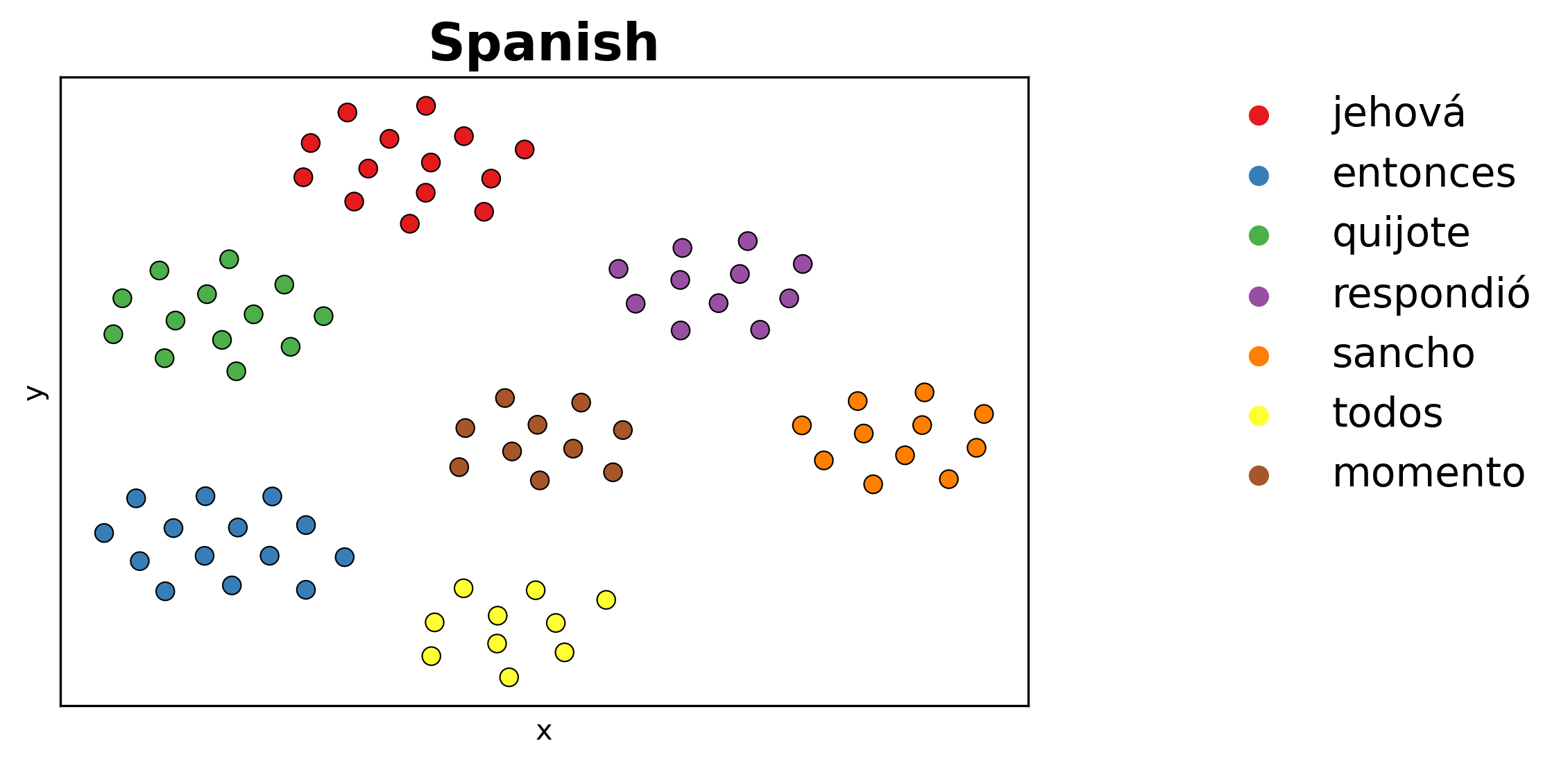}
    % \caption{Figure C}
    \label{fig:c}
  \end{subfigure}
  
  \begin{subfigure}[b]{\columnwidth}
    \centering
    \includegraphics[width=\columnwidth]{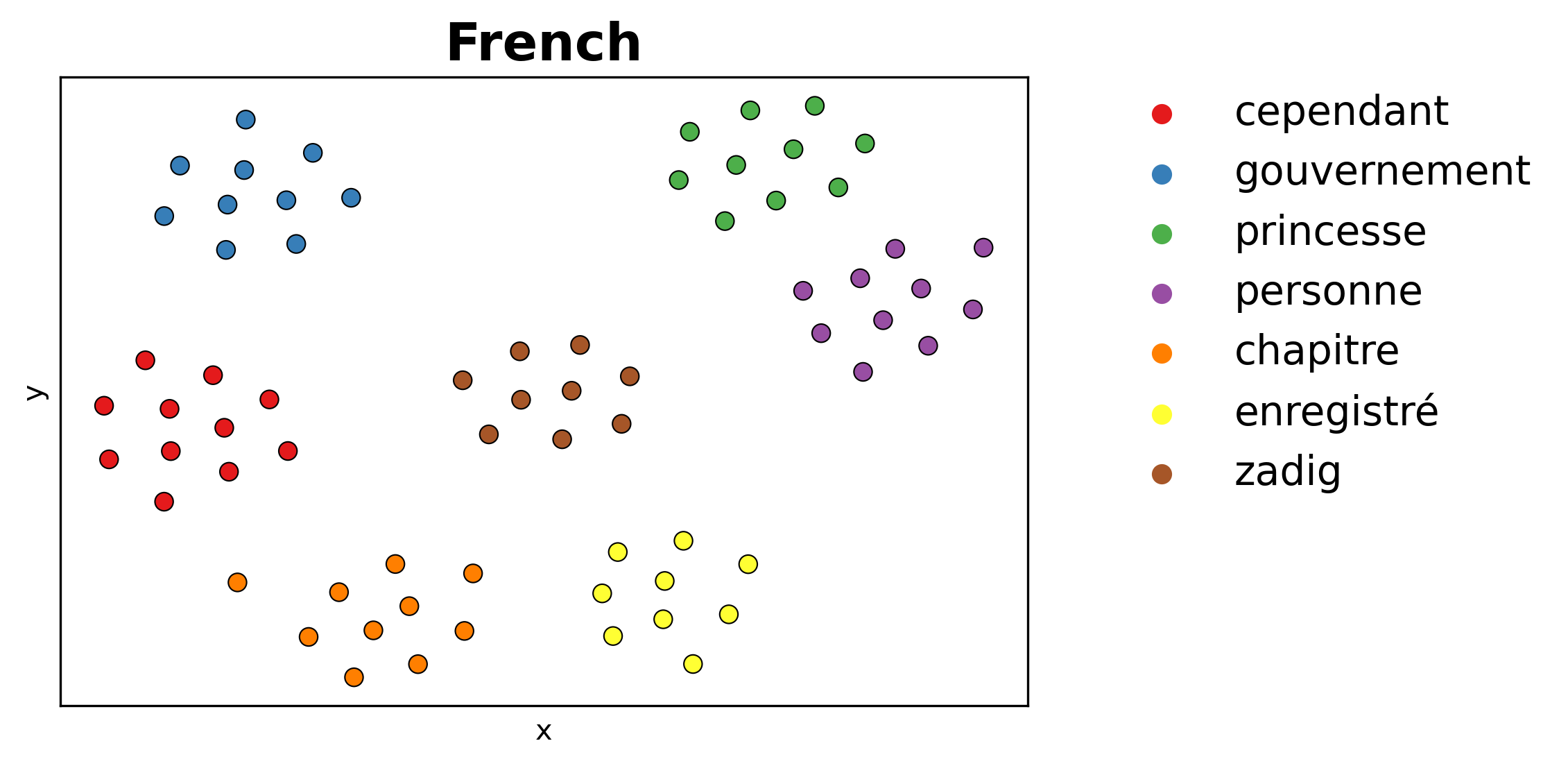}
    % \caption{Figure D}
    \label{fig:d}
  \end{subfigure}
  
  \begin{subfigure}[b]{\columnwidth}
    \centering
    \includegraphics[width=\columnwidth]{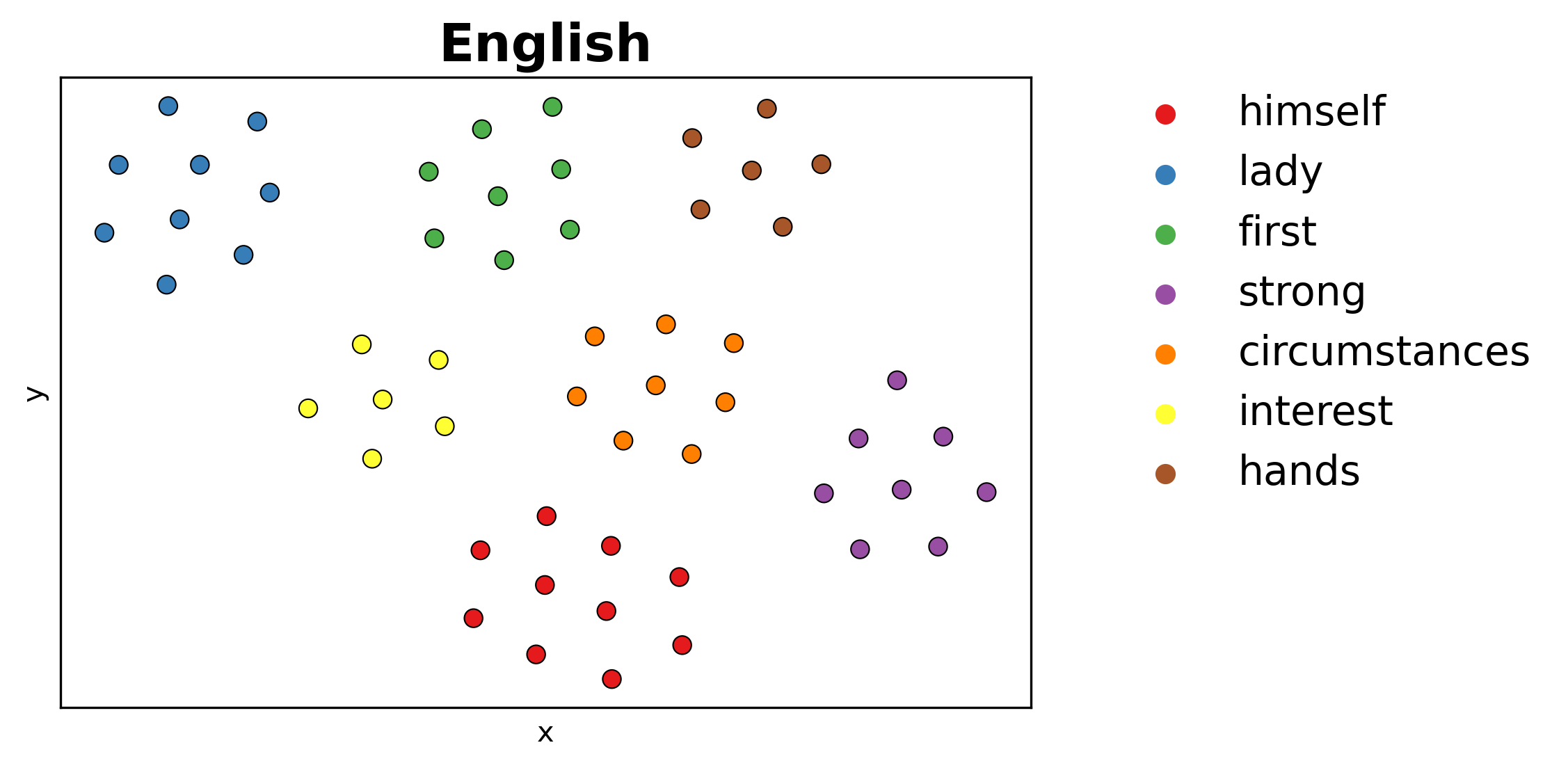}
    % \caption{Figure D}
    \label{fig:e}
  \end{subfigure}
  \caption{t-SNE visualisation of the AWEs derived from HuBERT-based CAE-RNN model for all five languages. From each language, all spoken instances of the top 7 words with
the highest frequency count from the test set are chosen.}
  \label{fig:four-figures}
\end{figure}

\subsection{Analysis of Anagram Pairs}
Anagrams are words that can be formed by rearranging the letters of another word. Analysing anagram pairs provides insights into the impact of letter order on AWE representation. Robust AWEs should capture the letter order in spoken words. For this analysis, same spoken word pairs and anagram pairs are chosen from different speakers. Ideally, the cosine distance between the same word pairs should be close to 0, while anagram word pairs should be close to 1. In Table \ref{tab7}, HuBERT-based CAE-RNN AWEs demonstrate cosine distances of approximately 0.01, 0.11, and 0.02 for the same spoken word pairs `aside', `this', and `no', respectively. The anagram pairs of the words `aside', `this', and `no' (i.e., `ideas', `hits', and `on') have distances of 0.99, 0.50, and 0.69, respectively, for the HuBERT-based CAE-RNN model. These values are significantly better for both the same word pairs and anagram word pairs when compared to the HuBERT-based mean pooling method \cite{AWE-icassp2023}. This indicates that the HuBERT-based CAE-RNN model accurately captures the letter order in a word compared to the mean pooling baseline \cite{AWE-icassp2023}.

\subsection{AWE Visualisation}
t-SNE visualization is used to plot the 2-dimensional representations of the derived AWEs for all five languages. From each language, all spoken instances of the top 7 words with the highest frequency count from the test set are chosen. The plots demonstrate distinct and well-separated clusters for each spoken word across all languages. One interesting pattern can be observed for the Polish language, where the clusters of the spoken words `Owadów' and `Owady' share the boundary and are closely related in the AWE space. This is likely due to the fact that the first four letters of both the words (o, w, a, d) are shared  and these words only differ in their endings.

\section{Conclusions and Future Work}
\label{sec:6}
It has been demonstrated that SSL-based speech representations with CAE-RNN models outperform mean pooling and AE-RNN models across all languages. They also outperform MFCC-based models. Among all the SSL models, HuBERT performs the best when used as input for the CAE-RNN model, outperforming models such as Wav2vec2 and WavLM. Notably, despite being pre-trained on English data, the SSL  models exhibit excellent performance on other languages, showcasing their cross-lingual generalization capability for AWE extraction.

Furthermore, quantitative analysis reveals that incorporating context information of the spoken word leads to more robust AWEs. The HuBERT-based CAE-RNN model trained on English language and tested on other target languages outperforms the mean pooling method and the CAE-RNN model trained on the target language using MFCC features. This `zero-shot' method to obtain robust AWEs for the target language can be useful in applications for low-resource languages \cite{hate_speech}. An analysis was also conducted to show that the CAE-RNN model effectively captures the order of letters in a word.

In future work, experiments will be conducted with the ``LARGE" variation of SSL models, as well as multilingual pre-trained SSL models such as Wav2vec2-XLSR \cite{wav2vec2-xlsr}. Additionally, an interesting experiment would involve training a single universal AWE model on all languages and comparing its performance with language-specific AWE models. Further research will focus on measuring the performance gains of SSL-based CAE-RNN models on downstream tasks such as query-by-example search \cite{QbE2,qbe_interspech18,ASE} and keyword spotting \cite{keyword-spotting}.

\section{Limitations}
This work is focused on the extraction of AWEs and measuring their quality solely based on the word discrimination task. No downstream applications such as query-by-example search and keyword spotting, have been discussed using the improved AWEs. In this work, only the ``BASE'' versions of the SSL-based speech models are explored for experiments and analysis. There are other variations, such as ``LARGE'' version, for which this study can be extended. All the languages considered in this work belong to the Indo-European language family. This work does not contain the analysis of languages that belong to another language family, such as Dravidian or Afroasiatic language families. This work does not deal with layer-wise analysis, which can provide better insights for further improving the AWEs.
\label{sec:7}

\section{Acknowledgments}
\label{sec:8}
This work was supported by the Centre for Doctoral Training in Speech and Language Technologies (SLT) and their Applications funded by UK Research and Innovation [grant number EP/S023062/1]. This work was also funded in part by LivePerson, Inc.

\bibliography{anthology,custom}

\end{document}